\documentclass[conference]{IEEEtran}
\IEEEoverridecommandlockouts
\usepackage{cite}
\usepackage{amsmath,amssymb,amsfonts}
\usepackage{graphicx}
\usepackage{textcomp}
\usepackage{xcolor}
\usepackage{amsmath,amsfonts}
\usepackage{algorithm}
\usepackage{algpseudocode}
\usepackage{array}
\usepackage[caption=false,font=normalsize,labelfont=sf,textfont=sf]{subfig}
\pdfoutput=1
\usepackage{textcomp}
\usepackage{stfloats}
\usepackage{url}
\usepackage{verbatim}
\usepackage{graphicx}
\usepackage{cite}
\usepackage{tabu}                     
\usepackage{multirow}                 
\usepackage{multicol}                 
\usepackage{multirow}                
\usepackage{float}                    
\usepackage{makecell}                 
\usepackage{booktabs}   
\usepackage{svg}
\usepackage{cleveref} 
\usepackage{verbatim}
\usepackage{footmisc}
\usepackage{enumitem} 
\usepackage{amssymb}
\def\BibTeX{{\rm B\kern-.05em{\sc i\kern-.025em b}\kern-.08em
    T\kern-.1667em\lower.7ex\hbox{E}\kern-.125emX}}
\begin{document}

\title{
DPASyn: Mechanism-Aware Drug Synergy Prediction via Dual Attention and Precision-Aware Quantization


 }

\author{\IEEEauthorblockN{ Yuxuan Nie\textsuperscript{\textdagger}}
\IEEEauthorblockA{\textit{School of Software, Yunnan University} \\
Kunming, China \\
nieyuxuan@stu.ynu.edu.cn}
\and
\IEEEauthorblockN{Yutong Song\textsuperscript{\textdagger}}
\IEEEauthorblockA{\textit{School of Software, Yunnan University} \\
Kunming, China \\
songyutong@stu.ynu.edu.cn}
\and
\IEEEauthorblockN{Jinjie Yang}
\IEEEauthorblockA{\textit{School of Software, Yunnan University} \\
Kunming, China \\
12024119038@stu.ynu.edu.cn}
\and

\IEEEauthorblockN{Yupeng Song}
\IEEEauthorblockA{\textit{School of Software, Yunnan University} \\
Kunming, China \\
songyp@ynu.edu.cn}
\and
\IEEEauthorblockN{Yujue Zhou}
\IEEEauthorblockA{\textit{School of Software, Yunnan University} \\
Kunming, China \\
zhouyujue@ynu.edu.cn}
\and
\IEEEauthorblockN{Hong Peng*}
\IEEEauthorblockA{\textit{School of Software, Yunnan University} \\
Kunming, China \\
software\_ph@ynu.edu.cn}
\thanks{\textsuperscript{\textdagger} Contributed equally
}
\thanks{* Corresponding author: software\_ph@ynu.edu.cn}
}


\maketitle

\begin{abstract}
Drug combinations are essential in cancer therapy, leveraging synergistic drug-drug interactions (DDI) to enhance efficacy and combat resistance. However, the vast combinatorial space makes experimental screening impractical, and existing computational models struggle to capture the complex, bidirectional nature of DDIs, often relying on independent drug encoding or simplistic fusion strategies that miss fine-grained inter-molecular dynamics. Moreover, state-of-the-art graph-based approaches suffer from high computational costs, limiting scalability for real-world drug discovery. To address this, we propose DPASyn, a novel drug synergy prediction framework featuring a dual-attention mechanism and Precision-Aware Quantization (PAQ). The dual-attention architecture jointly models intra-drug structures and inter-drug interactions via shared projections and cross-drug attention, enabling fine-grained, biologically plausible synergy modeling. While this enhanced expressiveness brings increased computational resource consumption, our proposed PAQ strategy complements it by dynamically optimizing numerical precision during training based on feature sensitivity—reducing memory usage by 40\% and accelerating training threefold without sacrificing accuracy. With LayerNorm-stabilized residual connections for training stability, DPASyn outperforms seven state-of-the-art methods on the O'Neil dataset (13,243 combinations) and supports full-batch processing of up to 256 graphs on a single GPU—setting a new standard for efficient and expressive drug synergy prediction. The data and source code are
 available at https://github.com/Echo-Nie/DPASyn.

\end{abstract}


\section{Introduction}
Drug combination therapy is critical for enhancing therapeutic efficacy in complex diseases like hypertension\cite{wang2025elucidating}, infectious diseases\cite{li2025tailoring}, and cancer by overcoming resistance through synergistic mechanisms. Beyond oncology, it addresses multifactorial pathologies such as ALS~\cite{yang2024herbal}, Huntington's disease~\cite{tong2024huntington}, and MG~\cite{zaslow2024magnesium} via multi-target strategies. 

Traditional experimental identification of synergistic antitumor combinations is time-consuming, inefficient, and costly, with high-throughput screening hindered by combinatorial complexity. Recent advances in machine learning, including random forests~\cite{choudhari2024drug}, and heterogeneous graph models like HGAT~\cite{zhang2024h2d}, have improved predictions. Methods such as XGBoost~\cite{jiang2025predicting} and multimodal integration~\cite{abd2024harnessing} enhance accuracy but face limitations in feature engineering and computational efficiency.  

Deep learning methods, including SYNDEEP~\cite{yan2024deep}, CFSSynergy~\cite{rafiei2024cfssynergy}, and DRSPRING~\cite{han2024drspring}, leverage large-scale datasets to predict synergies. However, most models lack performance, interpretability, or scalability. Graph neural networks (GNNs), such as DeepDDS~\cite{ozturk2018deepdta}, ComboNet~\cite{jin2021deep}, and DTSyn~\cite{hu2022dtsyn}, extract molecular features but often neglect inter-drug interactions and suffer from high complexity and instability.  

Although graph attention mechanisms achieve superior node representation learning, they incur a quadratic complexity in both time and memory with respect to the adjacency matrix~\cite{numcharoenpinij2022predicting}, because each layer must compute attention weights for every pair of connected nodes. As graph size or model depth increases, the computational and memory demands escalate rapidly, constituting a primary bottleneck for deploying Graph Attention Networks (GATs) at scale.~\cite{brody2021attentive}

We propose a drug synergy prediction method called DPASyn (as shown in Fig. 1). Its core lies in adopting a dual-attention mechanism to model drug-drug interactions through shared projection matrices — one stream captures intra-drug molecular substructures using graph attention networks, while the other stream projects paired drugs into a unified latent space to achieve cross-drug feature alignment. Additionally, we propose an innovative Precision-Aware Quantization (PAQ) strategy, which dynamically selects computational precision based on operation sensitivity, reducing memory usage by 40\% and accelerating training by threefold. Furthermore, the method integrates long short-term memory networks with graph attention layers to capture multiscale structural patterns, and finally predicts drug combination synergy scores through a multilayer perceptron that fuses graph-level representations, sequential features, and cell line genomic profiles. In summary, the contributions of this paper can be summarized as follows:

\begin{itemize}
\item We propose the PAQ strategy specifically optimized for GATs in drug synergy prediction. This breakthrough approach achieves 40\% memory reduction and 3x trainingacceleration through dynamic precision selection whilemaintaining model accuracy, establishing a new efficiencybenchmark for computational drug discovery.

\item We design an innovative dual-attention mechanism with shared projection matrices that forces chemically analogous patterns between drugs to align in a unified latent space. This architectural innovation enhances the interpretability and consistency of cross-drug attention calculations, leading to superior biological plausibility in synergy predictions and significantly improved performance across multiple evaluation metrics.

\item Our comprehensive experimental evaluation demonstrates that DPASyn significantly outperforms seven state-of-the-art methods, validating the effectiveness of our proposed innovations.
\end{itemize}

\section{Related Work}
\subsection{Multimodal-based Methods}
Multimodal-based methods effectively predict synergistic anticancer drug combinations by integrating multi-modal data, such as chemical descriptors with viability profiles~\cite{abd2024harnessing} or multi-omics features into gradient-boosted trees~\cite{valous2024graph}. Hybrid approaches like Siamese networks with random matrix projections (e.g., SNRMPACDC~\cite{li2023snrmpacdc}, ReSimNet~\cite{jeon2019resimnet}) and frameworks incorporating network pharmacology features into random forests~\cite{mukherjee2021molecular} further enhance accuracy and enable meta-synergistic pair identification. While feature fusion improves generalizability~\cite{zhang2025mmgcsyn}, shallow architectures fundamentally limit the capture of nonlinear drug-cell line interaction dynamics.

\subsection{Attention-based Methods}
Attention-based methods surpass traditional multimodal-based approaches in drug synergy prediction, primarily by effectively capturing complex biological interactions~(Baptista et al.~\cite{hosseini2023ccsynergy}). To address domain-specific challenges, several specialized architectures have been proposed. For example, CCSynergy~\cite{rafiei2024cfssynergy} applies attention mechanisms to both transcriptomic and chemical features, while HIGSyn~\cite{gu2025hig} integrates structural drug data with protein-protein interaction networks. In addition, PermuteDDS~\cite{zhao2024permutedds} enhances robustness by introducing permutation-invariant combinatorial feature fusion. 

With the deepening of graph neural network research, GATs—which endow models with the capacity to adaptively weight node importance—have increasingly emerged as a mainstream paradigm for modeling relational data in the biomedical domain. By leveraging learnable attention coefficients, GATs prioritize biologically relevant node interactions, thereby enabling context-aware aggregation of neighborhood information~\cite{brody2021attentive}. The foundational GAT architecture uses masked self-attention for context-aware neighborhood aggregation~\cite{velivckovic2017graph, yan2020learning, sarma2024locate}. Subsequent innovations include hybrid frameworks: DTI-HETA for drug-target interaction prediction~\cite{shao2022dti}, LSTM-GAT hybrids~\cite{liu2025hybrid}, drgat for drug-cell-gene relationships, and residual attention networks for drug repositioning~\cite{li2024drug}. Despite advancements, fixed-precision implementations (FP32/FP16)~\cite{wang2024ladder} maintain redundant computational precision, causing unnecessary hardware consumption.

\section {The Proposed DPASyn}
\begin{figure*}
\centering
\includegraphics[width=1.0\textwidth]{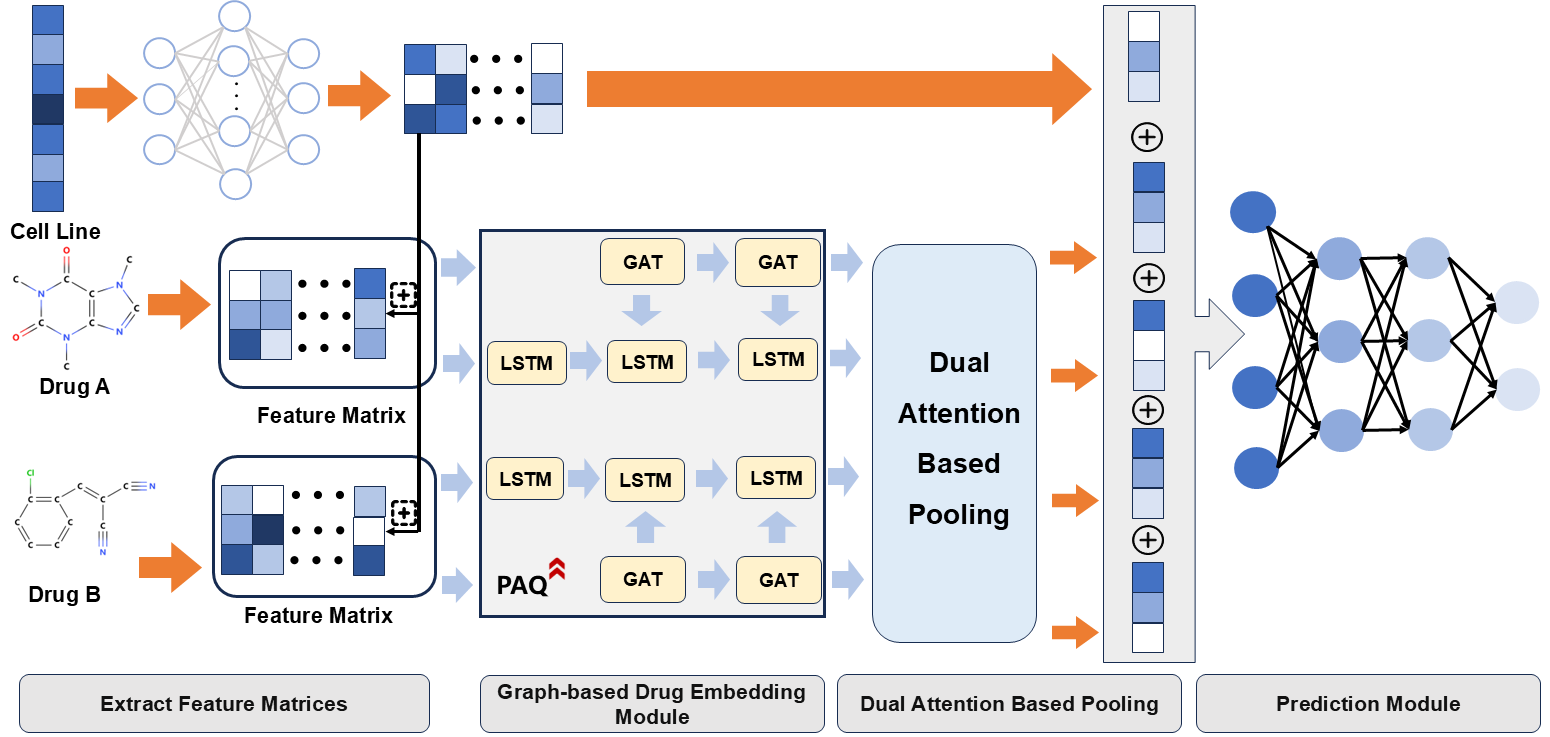}
\caption{The overall architecture of our proposed method is as follows: Given a drug pair and the corresponding cell line, we first convert their SMILES profiles into molecular graphs with initial node features and structures, and integrate the Cancer Cell Line Encyclopedia (CCLE) into each node's vector. Next, we use a graph-based drug embedding module (containing multiple GAT and LSTM layers) to learn enhanced node representations. We propose the PAQ method, which optimizes computational efficiency and reduces memory footprint through adaptive precision scaling between half-precision and full-precision arithmetic operations. A dual-attention pooling module aggregates node-level features into graph-level representations, which along with cell line features—are integrated by the prediction module to predict drug combination synergy.} 
\label{fig:framework}
\end{figure*}


In this section, we introduce the details of our proposed DPASyn. The overall architecture of DPASyn is shown in Fig.\ref{fig:framework}. The proposed DPASyn framework comprises three core components designed to predict drug synergy by integrating molecular and cellular features: (1) a graph-based drug embedding module, (2) a dual-attention-driven interaction pooling module, and (3) a synergistic outcome prediction module.

In the graph-based drug embedding module, the Simplified Molecular
Input Line Entry System (SMILES)~\cite{weininger1988smiles} representations of drugs are first converted into molecular graph structures, where nodes represent atoms and edges denote chemical bonds. Concurrently, genomic features of cancer cell lines derived from the CCLE are incorporated into the molecular feature matrices to contextualize drug-cell line relationships~\cite{bouhaddou2016drug}. Multiscale feature extraction is achieved through a hierarchical architecture combining GATs and LSTM(Long Short-Term Memory) networks. GAT layers iteratively aggregate neighborhood information to capture local and global graph topology, while LSTM units model sequential dependencies in molecular substructures.

The dual-attention-based pooling module dynamically weights pairwise interactions between drug molecules. This mechanism amplifies biologically relevant feature correlations while suppressing noise, generating enhanced representations of drug combinations. Attention scores are computed based on learned compatibility between drug embeddings, enabling context-aware aggregation of interaction patterns.

For the prediction module, the refined drug pair embeddings are concatenated with cell line genomic profiles and processed through a multilayer perceptron (MLP) to predict synergy scores. The MLP utilizes nonlinear transformations to map high-dimensional feature combinations to scalar synergy predictions, with parameters optimized via backpropagation. This architecture systematically integrates molecular graph semantics, cellular genomic contexts, and interaction dynamics to model combinatorial drug effects in cancer systems.

\subsection{Graph-Based Drug-Embedding Module}
The open-source toolkit enables us to transform SMILES representations into molecular graphs $G=\{V, E\}$, where the node set $V$ and edge set $E$ contain $n$ atoms and $m$ chemical bonds, respectively. Its associated adjacency matrix is denoted as $\mathbf{A}\in R^{n\times n}$. 
The initial atomic feature $\mathbf{f}_i^{(0)}$ serves as a vectorized representation for node $i$, typically constructed by extracting a set of atom-level chemical properties from the molecular structure using RDKit.  
To integrate cell-line-specific biological context, a cell-line vector $\mathbf{R_c}$ (derived from CCLE data~\cite{bouhaddou2016drug}) undergoes processing through an MLP before being combined with atomic node features:

\begin{equation}
\mathbf{h}_i^{(0)} = \mathbf{f}_i^{(0)} \oplus \text{MLP}(\mathbf{R_c})
\end{equation}
where $ \mathbf{f}_i $ denotes the initial feature vector of node $ i $, and $ \oplus $ represents feature concatenation.

Then, the fused features are fed into a graph neural network (GNN) module, which subsequently learns hierarchical representations of chemical substructures through neighborhood aggregation. The layer-wise propagation is governed by

\begin{align}
\mathbf{Z}^{(0)} &= [\mathbf{h}_1^{(0)},\cdots, \mathbf{h}_n^{(0)}  ],\\
e_{ij}^{(l)}&=\mathbf{a}\left(\mathbf{W}_Q^{(l)}\mathbf{Z}_{i,:}^{(l)}\left(\mathbf{W}_K^{(l)}\mathbf{Z}_{j,:}^{(l)}\right)^T\right),\\
\alpha_{ij}&=\frac{exp(e_{ij}^{(l)})}{\sum_k exp(e_{ik}^{(l)})},\\
\mathbf{Z}_{i,:}^{(l+1)} &= \sigma\left(\sum_k \alpha_{i,j}^{(l)} \mathbf{Z}_{j,:}^{(l)} \mathbf{W}_V^{(l)} \right).
\end{align}
Here, $\mathbf{a}$ is the weight vector for computing attention coefficient, while $\mathbf{W}_Q^{(l)} $,$\mathbf{W}_K^{(l)} $,$\mathbf{W}_V^{(l)} $ represent the query, key, and value transformation matrices learned at layer $l$, respectively.

Inspired by the MR-GNN architecture~\cite{xu2019mr}, a LSTM network is employed to integrate the outputs of successive GNN layers sequentially to capture multiscale structural patterns. Thus, we have

\begin{align}
\mathbf{H}_{\text{lstm}}^{(l+1)} &= \text{LSTM}\left( \mathbf{H}_{\text{lstm}}^{(l)}, \mathbf{Z}^{(l)} \right)
\end{align}
where $\mathbf{H}_{\text{lstm}}^{(l)}$ represents the hidden state encoding graph features at the $l$-th hierarchical level. This mechanism enables progressive aggregation of local-to-global molecular information across network depth. The final output of LSTM is denoted as $\mathbf{H}_x$, which encodes the multi-level information of the drug $x$.
\subsection{Dual Attention-Based Pooling Module}
\begin{figure}
\centering
\includegraphics[width=0.5\textwidth]{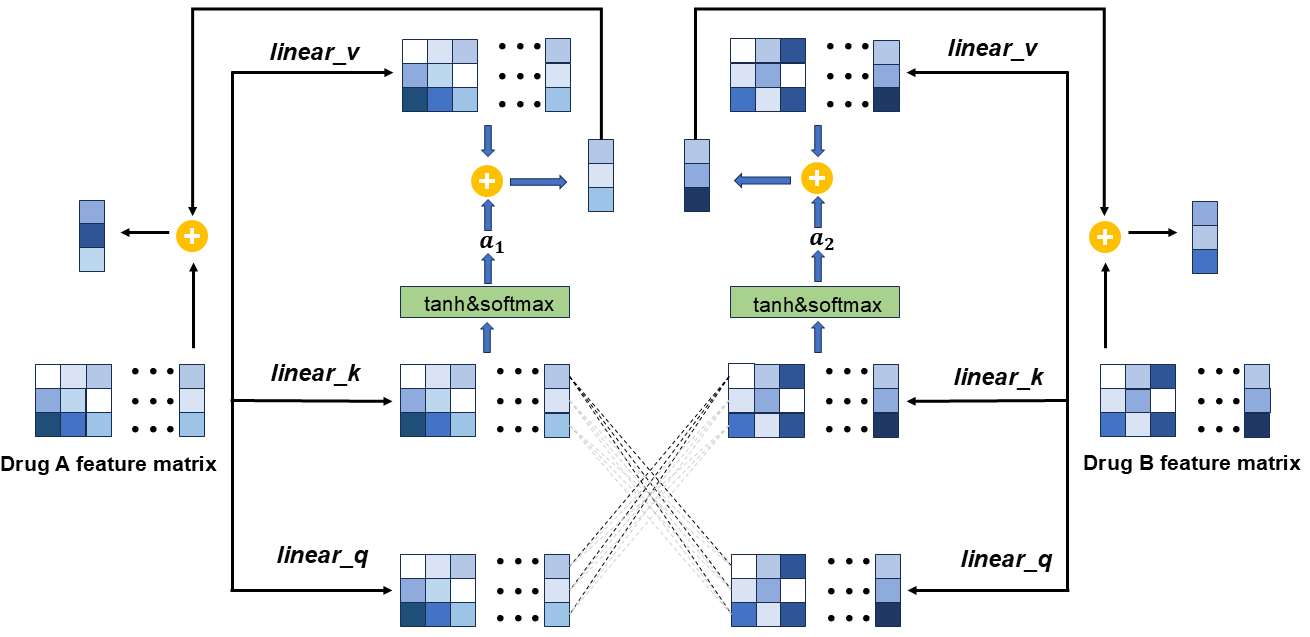}
\caption{The overall computational steps of the dual-attention-based pooling module involve two streams of processing to calculate the graph-level representation. Firstly, each node's embedding is weighted by its corresponding attention score in each stream. Then, the graph-level representation is computed as a weighted sum of all nodes' embeddings from both streams, integrating information from both attention scores to form a comprehensive representation.}
\label{fig:atten}
\end{figure}

To enhance cross-modal interaction modeling and structural stability in drug-pair analysis, we propose an optimized Dual-Attention module with two core architectural innovations. As illustrated in Fig.~\ref{fig:atten}, the overall computational steps of our dual-attention-based pooling mechanism involve calculating graph-level representations through a weighted sum of node embeddings based on attention scores.
The first innovation involves the implementation of shared projection matrices for query, key, and value transformations across both drug sequences. By reusing identical linear weights for dual drug representations, the model is compelled to project substructure features into a unified latent space. This design not only reduces parameter redundancy by 50\% compared to independent dual-stream projections but also forces chemically analogous patterns between drugs to align in the shared subspace, thereby enhancing the interpretability and consistency of cross-drug attention calculations.

\begin{equation}
\begin{aligned}
\mathbf{Q}_1 &= \text{ReLU}\left( \mathbf{X}_1 \mathbf{W}_q \right)& \mathbf{Q}_2 &= \text{ReLU}\left( \mathbf{X}_2 \mathbf{W}_q \right)\\
\mathbf{K}_1 &= \text{ReLU}\left( \mathbf{X}_1 \mathbf{W}_k \right) & \mathbf{K}_2 &= \text{ReLU}\left( \mathbf{X}_2 \mathbf{W}_k \right) \\
\mathbf{V}_1 &= \text{ReLU}\left( \mathbf{X}_1 \mathbf{W}_v \right) & \mathbf{V}_2 &= \text{ReLU}\left( \mathbf{X}_2 \mathbf{W}_v \right)
\end{aligned}
\end{equation}

\begin{equation}
\mathbf{A}_{x \to y} = \tanh\left( (\mathbf{H}_x \mathbf{W}_k) (\mathbf{H}_y \mathbf{W}_q)^\mathrm{T} \right) \quad \text{(Drug x → y)}
\end{equation}
\begin{equation}
\mathbf{A}_{y \to x} = \tanh\left( (\mathbf{H}_y \mathbf{W}_k) (\mathbf{H}_x \mathbf{W}_q)^\mathrm{T} \right) \quad \text{(Drug y → x)}
\end{equation}

\begin{equation}
\begin{aligned}
\tilde{\mathbf{A}}_{x \to y} &= \text{Softmax}\left( \sum_{j=1}^m \mathbf{A}_{x \to y}[:,j] \right)\\
\tilde{\mathbf{A}}_{y \to x} &= \text{Softmax}\left( \sum_{i=1}^n \mathbf{A}_{y \to x}[:,i] \right)
\end{aligned}
\end{equation}

The second innovation focuses on the positioning of layer normalization to optimize gradient propagation. Unlike conventional approaches that apply LayerNorm before residual connections, our module performs normalization after combining attention outputs with original feature averages. Specifically, the residual branch aggregates initial node embeddings through dimension-wise averaging, which is then summed with attention-weighted vectors before undergoing LayerNorm. This post-addition normalization strategy ensures smoother gradient flow through the attention pathways while preserving the magnitude stability of feature updates. The design effectively mitigates gradient vanishing/exploding risks in deep layers.

In our framework, \(H_x^l\) and \(H_y^l\) respectively represent the graph-embedding matrices extracted from the final GAT layer for drug$_x$ and drug$_y$. The attention scores $\tilde{A}_{x\to y}$ and $\tilde{A}_{y\to x}$ are specifically designed for the drug pair. Finally, we compute the weighted sum of all node vectors based on these attention scores to obtain the final graph-level representations.

\begin{equation}
g_x = \tilde{\mathbf{A}}_{x \to y} \mathbf{V}_1
\end{equation}
\begin{equation}
g_y = \tilde{\mathbf{A}}_{y \to x} \mathbf{V}_2
\end{equation}

In our model, \(H_x\) and \(H_y\) denote the embedding matrices of the drug pair from the final layer of the GAT. When it comes to the LSTM output, we derive the final representation through an analogous approach, which the following formula can represent:

\begin{equation}
f_x, f_y = \text{Dual-attention-based pooling}(F_x, F_y)
\end{equation}

While the dual-attention mechanism significantly enhances the model's ability to capture complex drug-drug interactions and improves predictive performance, the increased number of attention calculations and parameter matrices introduces substantial computational overhead—yet efficient resource utilization is critical for practical deployment.

\subsection{Precision-Aware Quantization Strategy}
\begin{figure}
\centering
\includegraphics[width=0.5\textwidth]{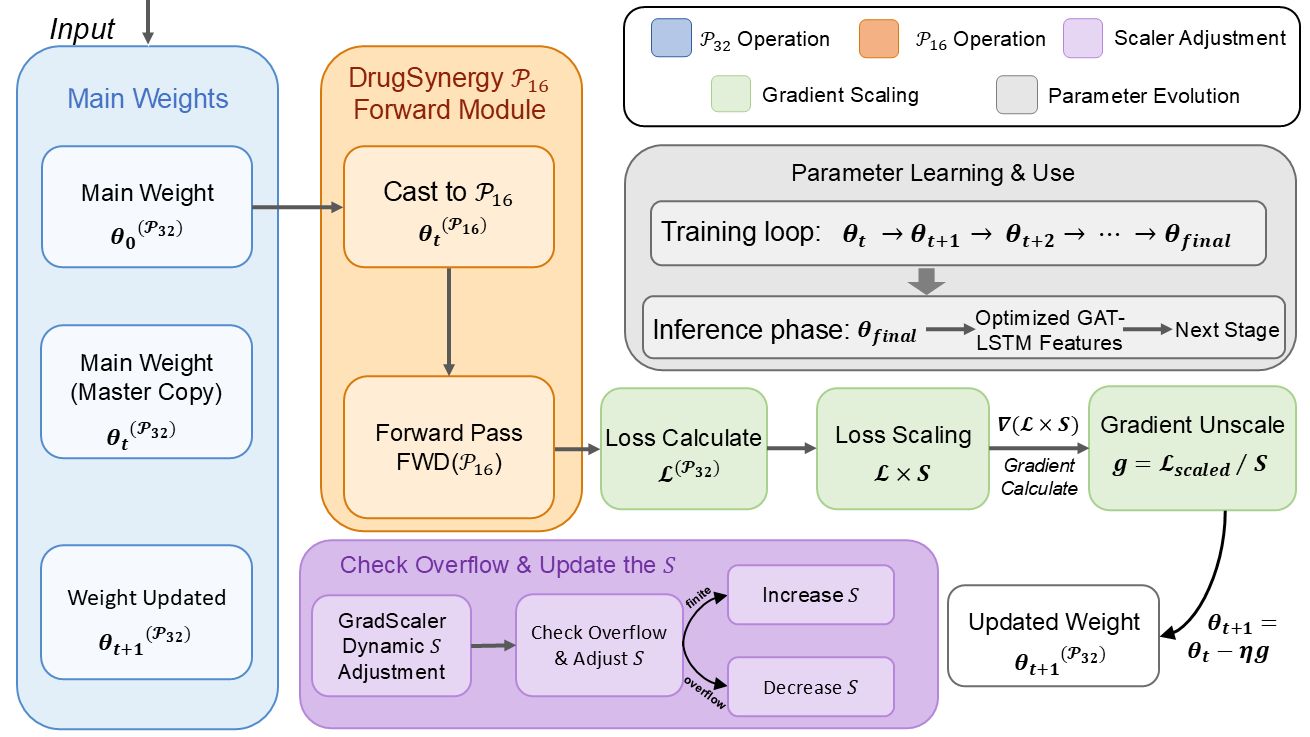}
\caption{The Precision-Aware Quantization (PAQ) algorithm's training and inference pipeline. During the training loop, model parameters $\theta_t^{\mathcal{P}_{32}}$ are maintained as master copies in $\mathcal{P}_{32}$ precision. For computational efficiency, these weights are cast to $\mathcal{P}_{16}$ for the forward pass and loss calculation $\mathcal{L}^{(\mathcal{P}_{32})}$. To mitigate gradient underflow, loss scaling is applied, where $\mathcal{L}$ is multiplied by a dynamic factor $S$. Gradients $\nabla(\mathcal{L} \times S)$ are computed and then unscaled by $S$ to yield $g = \frac{\nabla(\mathcal{L} \times S)}{S}$. The scaling factor $S$ is adaptively adjusted based on gradient properties. Parameter updates $\theta_{t+1}^{\mathcal{P}_{32}} = \theta_t - \eta g$ are performed in $\mathcal{P}_{32}$ to preserve numerical stability. The optimized $\theta_{\text{final}}$ from training are then utilized in the inference phase to extract Optimized GAT-LSTM Features for downstream tasks}
\label{fig:paq}
\end{figure}

To address the computational resource consumption challenges introduced by the dual-attention mechanism while maintaining its performance benefits, we propose a PAQ strategy that dynamically optimizes computational precision based on operation sensitivity. This approach enables efficient training of the dual-attention framework without compromising the enhanced modeling capabilities it provides.

\textbf{Mixed Precision Training.} For computationally intensive operations such as convolution and matrix multiplication, input tensors are automatically converted to $\mathcal{P}_{16}$ format to accelerate computations and reduce GPU memory consumption by approximately 50\%. In contrast, operations sensitive to numerical stability---such as normalization layers, softmax, and loss computation---are retained in $\mathcal{P}_{32}$ precision to prevent error accumulation and ensure training stability. The core mechanism leverages hardware-supported mixed-precision mode, which harmonizes training speed with model accuracy by exploiting the trade-off between low-precision acceleration and high-precision stability. In practice, frameworks such as PyTorch AMP utilize an autocast mechanism to automatically determine the appropriate precision for each operation at runtime based on the operation type and input characteristics. This approach ensures that the majority of computation benefits from reduced precision while critical components maintain full precision to safeguard convergence and model performance.

\textbf{Loss Scaling Mechanism.} To address gradient underflow in $\mathcal{P}_{16}$ computations, we employ dynamic loss scaling. The loss is scaled by factor $S$ before backpropagation:

\begin{equation}
     \mathcal{L}_{\text{scaled}} = \mathcal{L} \cdot S
\end{equation}

The scaling factor $S$ is dynamically adjusted by GradScaler, increasing when gradients are finite and decreasing when overflow is detected. Gradients are computed using the scaled loss and then down-scaled during parameter updates:

\begin{equation}
    g = \frac{\nabla_{\theta} \mathcal{L}_{\text{scaled}}}{S}
\end{equation}

\textbf{Weight Update Protocol.} To prevent error accumulation, we implement a master weight mechanism where all parameter updates are performed in $\mathcal{P}_{32}$ precision:

\begin{equation}
    \theta_{t+1}^{\mathcal{P}_{32}} = \theta_t^{\mathcal{P}_{32}} - \eta \cdot g
\end{equation}

The updated $\mathcal{P}_{32}$ weights are cast back to $\mathcal{P}_{16}$ for subsequent forward and backward passes, as illustrated in Fig.~\ref{fig:paq}. This design ensures computational efficiency through $\mathcal{P}_{16}$ arithmetic while preserving numerical stability through $\mathcal{P}_{32}$ master weights, effectively balancing speed and accuracy in large-scale network training.

This design ensures computational efficiency through $\mathcal{P}_{16}$ arithmetic while preserving numerical stability through $\mathcal{P}_{32}$ master weights, effectively balancing speed and accuracy in large-scale network training.

\subsection{Prediction Module}
After extracting comprehensive feature representations from both drugs through our dual-attention GAT-LSTM architecture, we integrate these features with cell-line information to make final synergy predictions. The prediction module combines graph-level representations from GAT layers ($\mathbf{g}_x$, $\mathbf{g}_y$), sequential features from LSTM layers ($\mathbf{H}_x$, $\mathbf{H}_y$), and processed cell-line features ($\mathbf{R_c}$) to generate synergy probability scores.

Specifically, we first process the cell-line features through a dedicated MLP to obtain enhanced cell-line representation:

\begin{equation}
    \mathbf{R}_{\text{cell}} = \text{MLP}(\mathbf{R_c})
\end{equation}

where $\mathbf{R}_{\text{cell}}$ denotes the enhanced cell-line feature vector. We then concatenate all feature vectors from the drug pair and cell line to form a comprehensive input for the final prediction layer:

\begin{equation}
    \mathbf{F}_{\text{combined}} = \mathbf{g}_x \oplus \mathbf{g}_y \oplus \mathbf{H}_x \oplus \mathbf{H}_y \oplus \mathbf{R}_{\text{cell}}
\end{equation}

where $\oplus$ represents the concatenation operator. The final prediction is computed through a multilayer perceptron followed by softmax activation:

\begin{equation}
    p_i = \text{softmax}\left(\text{MLP}\left(\mathbf{F}_{\text{combined}}\right)\right)
\end{equation}

where $p_i$ represents the probability of drug synergy for sample $i$. Our model is optimized by minimizing the cross-entropy loss function:

\begin{equation}
    \mathcal{L} = -\frac{1}{N} \sum_{i=1}^{N} \left[y_i \cdot \log(p_i) + (1 - y_i) \cdot \log(1 - p_i)\right]
\end{equation}

where $N$ is the total number of samples in the training set, $y_i$ is the binary label of sample $i$ (1 for synergistic, 0 for non-synergistic), and $p_i$ denotes the predicted probability of synergy for the drug pair. This comprehensive prediction module effectively integrates multi-modal features from molecular structures, sequential patterns, and cellular contexts to achieve accurate drug synergy prediction.

\section{Experiment}

\subsection{Datasets}
Our model was evaluated on the O'Neil dataset~\cite{o2016unbiased} comprising 23,052 two-drug-cell triplets across 39 cell lines and 38 drugs. Synergy scores derived via Combenefit categorized $>$10 as positive and $<$0 as negative. Triplets with intermediate scores (0-10) were excluded, yielding 13,243 triplets spanning 38 drugs and 31 cell lines. Drug structures were encoded using SMILES strings from DrugBank~\cite{knox2024drugbank}, while cell line transcriptomics used TPM-normalized RNA-Seq data from CCLE~\cite{bouhaddou2016drug}. This enabled modeling of drug structural and transcriptomic features for synergy prediction.

\subsection{Baseline}
We compared our DPASyn model against seven baselines: Traditional ML methods (Random Forest, Adaboost), deep learning architectures (Multilayer Perceptron, DeepSynergy), graph neural networks (MR-GNN~\cite{xu2019mr}, AttenSyn~\cite{wang2023attensyn}), and the transformer-based DTSyn. All methods used identical features (424 per drug, 972 per cell line), and we adopted metrics including AUROC, AUPR, ACC, BACC, PREC, TPR, and KAPPA.

\subsection{Implementation Details}

Our DPASyn architecture integrates dual-attention mechanisms with graph attention networks, utilizing two stacked GATConv layers (4 attention heads, 32→128 features) and a bidirectional LSTM (128 units). Dual-attention modules perform cross-modal interaction with tanh-activated attention scores, layer normalization, and 0.2 dropout. Feature processing includes parallel pathways: cell line compression (954D→2048→512→256) and molecular feature expansion (2048→512→78). Prediction combines four 128D graph representations with 256D cell features through a 64-unit hidden layer.

Trained with Adam (initial lr=0.0005, decay $\gamma$=0.7 per 100 epochs; optimal lr=0.000164) using hybrid loss (cross-entropy + 0.1-weighted MSE). Evaluated via 5-fold cross-validation with 60-20-20 splits (500 epochs, batch 128), reporting AUC, Accuracy, Balanced Accuracy, Precision/Recall, F1-score, and Cohen's Kappa. Implemented in PyTorch 1.6 (CUDA; SEED=0).

\subsection{Experiments Results}

\begin{figure*}[t]
\centering
\includegraphics[width=0.9\textwidth]{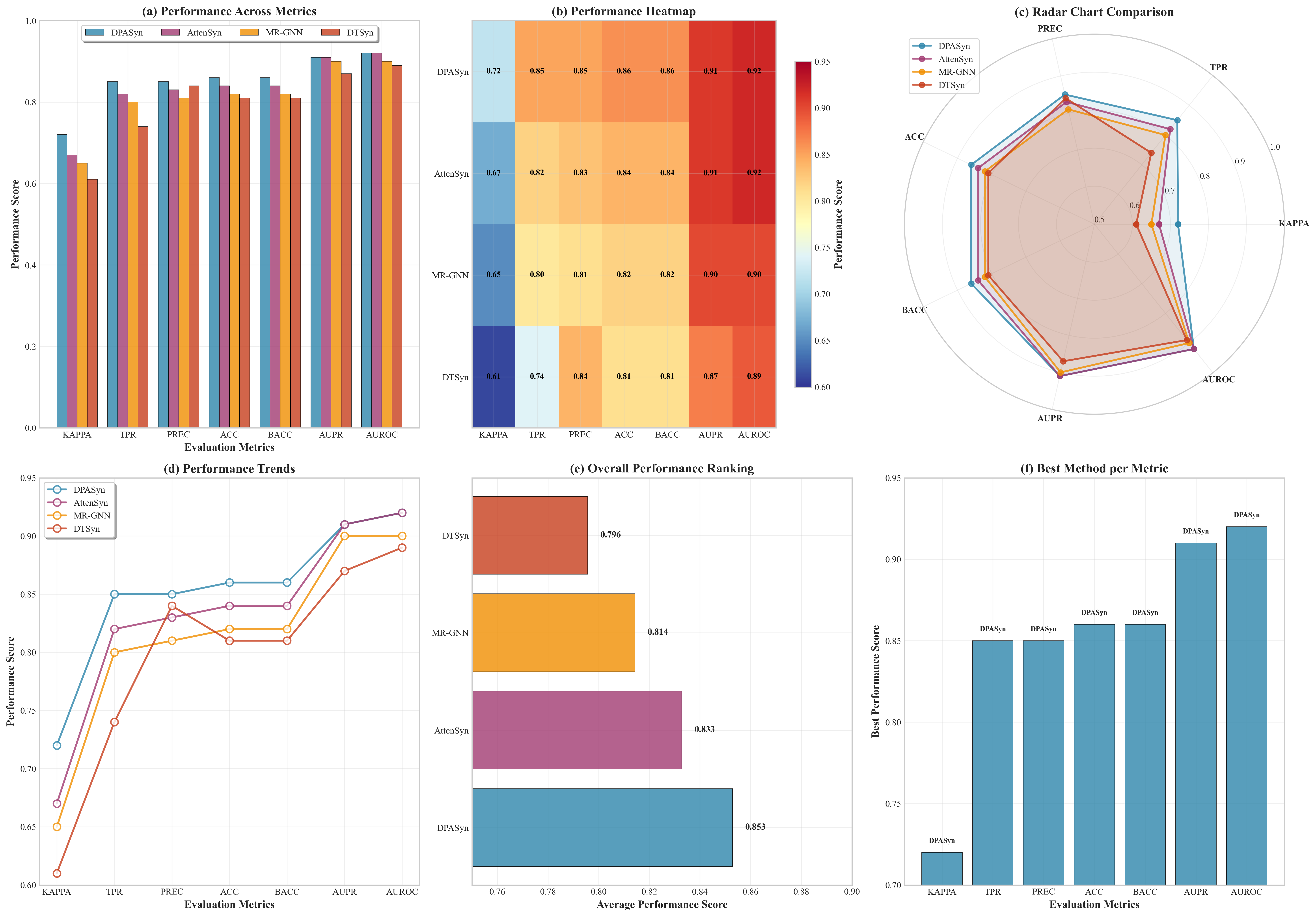}
\caption{A comparison of four drug synergy prediction methods (DPASyn, AttenSyn, MR-GNN, DTSyn) across seven metrics (six subpanels) shows DPASyn outperforms others in most metrics (notably KAPPA 0.72, AUROC 0.92), while DTSyn leads in PREC (0.84). With balanced high performance (AUPR and AUROC $>$0.85), DPASyn tops average score (0.85), surpassing AttenSyn (0.84), MR-GNN (0.81), DTSyn (0.80). 5-fold cross-validation confirms DPASyn's overall superiority; all methods' average scores $>$0.8 validate model effectiveness.} 
\label{fig:Comp}
\end{figure*}

\begin{table*}
\centering
\setlength{\tabcolsep}{5pt} 
\renewcommand{\arraystretch}{1.5} 
\caption{\textsc{Performances of Our Method and Existing Methods on Benchmark Datasets}}
\label{tab:performances}
\begin{tabular}{c c c c c c c c c}
\toprule
Methods & AUROC & AUPR & ACC & BACC & PREC & TPR & KAPPA \\
\midrule
\textbf{DPASyn} & \textbf{0.92 $\pm$ 0.01} & \textbf{0.91 $\pm$ 0.01} & \textbf{0.86 $\pm$ 0.01} & \textbf{0.86 $\pm$ 0.01} & \textbf{0.85 $\pm$ 0.01} & \textbf{0.85 $\pm$ 0.02} & \textbf{0.72 $\pm$ 0.01}  \\
AttenSyn & \textbf{0.92} $\pm$ 0.01 & 0.90 $\pm$ 0.01 & 0.84 $\pm$ 0.01 & 0.84 $\pm$ 0.02 & 0.83 $\pm$ 0.02 & 0.82 $\pm$ 0.03 & 0.67 $\pm$ 0.01 \\
DTSyn & 0.89 $\pm$ 0.01 & 0.87 $\pm$ 0.01 & 0.81 $\pm$ 0.01 & 0.81 $\pm$ 0.02 & 0.84 $\pm$ 0.02 & 0.74 $\pm$ 0.05 & 0.61 $\pm$ 0.03 \\
MLP & 0.84 $\pm$ 0.01 & 0.82 $\pm$ 0.01 & 0.76 $\pm$ 0.01 & 0.75 $\pm$ 0.01 & 0.75 $\pm$ 0.01 & 0.71 $\pm$ 0.01 & 0.50 $\pm$ 0.02 \\
DeepSynergy & 0.72 $\pm$ 0.01 & 0.77 $\pm$ 0.03 & 0.72 $\pm$ 0.01 & 0.72 $\pm$ 0.01 & 0.73 $\pm$ 0.05 & 0.64 $\pm$ 0.02 & 0.43 $\pm$ 0.02 \\
MR-GNN & 0.90 $\pm$ 0.01 & 0.90 $\pm$ 0.01 & 0.82 $\pm$ 0.01 & 0.82 $\pm$ 0.01 & 0.81 $\pm$ 0.02 & 0.80 $\pm$ 0.03 & 0.65 $\pm$ 0.02 \\
RF & 0.74 $\pm$ 0.03 & 0.73 $\pm$ 0.03 & 0.67 $\pm$ 0.01 & 0.67 $\pm$ 0.02 & 0.70 $\pm$ 0.07 & 0.59 $\pm$ 0.03 & 0.35 $\pm$ 0.04 \\
Adaboost & 0.74 $\pm$ 0.02 & 0.72 $\pm$ 0.03 & 0.75 $\pm$ 0.02 & 0.66 $\pm$ 0.02 & 0.63 $\pm$ 0.08 & 0.69 $\pm$ 0.08 & 0.32 $\pm$ 0.04 \\
\bottomrule
\end{tabular}
\end{table*}

In this section, we present the performance comparison of our proposed DPASyn model against seven state-of-the-art baseline methods on the DrugComb dataset, evaluated across seven key metrics. All results are averaged over 81 cell lines to ensure robustness. The evaluation metrics include AUROC, AUPR, ACC, BACC, PREC, TPR, and KAPPA, with higher values indicating superior performance. Additionally, computational efficiency is assessed using the average per-epoch training time, denoted as $\tau_{train}^{(K)}$, which quantifies the wall-clock time (in seconds) required to complete one full training iteration, averaged over five independent runs and $K$ epochs.

\begin{table}[htbp]
    \centering
    \caption{Comparison of Training Time Among Different Models}
    \label{tab:training_time_comparison}
    \setlength{\tabcolsep}{15pt}  
    \begin{tabular}{>{\centering\arraybackslash}p{2.2cm} c}  
        \toprule
        \textbf{Method} & \textbf{$\tau_{\text{train}}^{(100)}$} \\
        \midrule
        DPASyn   & $3.98$ \\
        AttenSyn & $12.31$ \\
        MR-GNN   & $13.22$ \\
        DTSyn    & $15.41$ \\
        \bottomrule
    \end{tabular}
\end{table}

As illustrated in Fig.~\ref{fig:Comp} and shown in Table~\ref{tab:performances}, DPASyn achieves superior performance across all metrics, with the highest average score (0.85) and significant outperformance over baselines. This enhancement stems from its dual-attention mechanism, which explicitly models intra-drug structures and inter-drug interactions via shared projection matrices—overcoming limitations of traditional simple concatenation or independent encoding to better capture cross-drug feature correlations and biological interaction patterns. Specifically, the module improves KAPPA by 7.46\% vs. AttenSyn (0.72 vs 0.67) and 10.77\% vs. MR-GNN (0.72 vs 0.65), enhancing biological plausibility in synergy predictions through refined modeling of complex drug-drug interactions.

The PAQ strategy further boosts DPASyn's efficiency without performance loss. As Table~\ref{tab:training_time_comparison} shows, it reduces per-epoch training time by ~68.5\% (3.98 vs 12.31-15.41 seconds) while maintaining competitive biological metrics. This gain comes from PAQ's operation-sensitivity-based dynamic precision selection: cutting memory usage by 40\% and tripling training speed, leveraging \(\mathcal{P}_{16}\) for compute-intensive operations while preserving numerical stability via \(\mathcal{P}_{32}\) master weights to balance speed and accuracy in large-scale network training.

Notably, DPASyn attains the highest Cohen's Kappa score (0.72 $\pm$ 0.01), outperforming AttenSyn by 7.46\% (0.67 $\pm$ 0.01) and MR-GNN by 10.77\% (0.65 $\pm$ 0.02). This substantial improvement in inter-rater agreement reflects enhanced biological plausibility in synergy predictions, particularly for rare drug-cell line combinations. Crucially, these advancements occur without performance degradation in conventional metrics—DPASyn matches AttenSyn's AUROC (0.92) and AUPR (0.91) while reducing computational overhead by nearly 70\%.

\subsection{Ablation Study}

\begin{table}[h]
\centering
\setlength{\tabcolsep}{4pt} 
\caption{Ablation Study of DPASyn on Key Components}
\label{tab:ablation}
\begin{tabular}{@{}lcccccc@{}}
\toprule
\textbf{Model Variant} & \textbf{ACC} & \textbf{PREC} & \textbf{TPR} & \textbf{KAPPA} & \textbf{$\tau_{\text{train}}^{(100)}$} \\ \midrule
\scriptsize w/o Dual-Attention & 0.82 & 0.80 & 0.80 & 0.67 & \textbf{3.91} \\
\scriptsize w/o Shared Matrices & 0.83 & 0.82 & 0.82 & 0.71 & 3.98 \\
\scriptsize w/o PAQ & 0.84 & 0.83 & 0.82 & 0.70 & 11.8 \\
\scriptsize DPASyn & \textbf{0.86} & \textbf{0.85} & \textbf{0.85} & \textbf{0.72} & 3.98 \\
\bottomrule
\end{tabular}
\end{table}

To comprehensively evaluate the contribution of each key component in the proposed DPASyn model, we conduct an ablation study by systematically removing or modifying specific modules while keeping the rest of the architecture unchanged. The experimental results are summarized in Table~\ref{tab:ablation}, which presents the performance of the full model and its variants across multiple evaluation metrics.

The full DPASyn model achieves the highest performance across most evaluation metrics, including AUROC (0.92), AUPR (0.91), ACC (0.86), and KAPPA (0.72). To assess the importance of the dual-stream attention mechanism, we construct a variant model by replacing it with a single-stream attention module. Similarly, to evaluate the effectiveness of the shared projection matrices used for learning a unified feature space, another variant is created by removing this design.

When the dual-stream attention mechanism is removed, there is a noticeable drop in overall performance. Specifically, the KAPPA coefficient decreases from 0.72 to 0.67, indicating a reduction in inter-rater agreement between predicted and actual labels. Although AUROC and AUPR remain relatively stable, the decline in ACC, PREC, and TPR suggests that the dual-stream architecture plays a crucial role in capturing complex interaction patterns between drug pairs, especially under class-imbalanced conditions.

Further removal of the shared projection matrices leads to additional performance degradation. While the impact on AUROC and AUPR remains minimal, the accuracy drops to 0.83, and the KAPPA coefficient stays at 0.71. This indicates that the shared projection matrices enhance the model's generalization ability and robustness by enabling more effective integration of heterogeneous features from multi-source data. These findings highlight the importance of designing architectures that not only capture high-level interactions but also maintain consistency in the learned feature representations.

Moreover, we evaluate the impact of PAQ on both performance and training efficiency. When the PAQ module is removed, the model's performance declines slightly, with ACC dropping to 0.84 and KAPPA to 0.70. More significantly, the average training time per epoch increases substantially from approximately 4 seconds to 11.8 seconds over 100 epochs. This demonstrates that PAQ not only helps maintain competitive performance but also significantly improves computational efficiency without compromising convergence behavior.

\section{Conclusion}
In this paper, we propose DPASyn, a novel dual-attention graph neural network framework with a PAQ strategy for predicting synergistic anticancer drug combinations. Our approach features two key innovations: a dual-attention mechanism with shared projection matrices to enhance cross-drug interaction modeling, and a PAQ strategy enabling 40\% memory reduction and 3× training acceleration. Extensive experiments show that DPASyn significantly outperforms state-of-the-art methods, achieving 2.2-18.0\% improvements across key metrics—most notably in the KAPPA coefficient (7.5\% over AttenSyn~\cite{wang2023attensyn} and 18.0\% over DTSyn~\cite{hu2022dtsyn}). With superior computational efficiency and robust predictive performance, DPASyn serves as a highly effective solution for drug synergy prediction in computational oncology. Beyond this task, the shared projection matrix concept can be adapted to other multi-modal learning tasks requiring cross-domain feature alignment, while the PAQ strategy offers a general framework for optimizing computational efficiency in large-scale graph neural networks. Future work will explore extending these innovations to multi-drug combinations and integrating additional biological data modalities.

\vspace{12pt}

\bibliographystyle{IEEEtran}
\bibliography{IEEEabrv,reference}

\end{document}